# IF-Net: An Illumination-invariant Feature Network

Po-Heng Chen, Zhao-Xu Luo, Zu-Kuan Huang, Chun Yang, Kuan-Wen Chen*

*Abstract*— Feature descriptor matching is a critical step is many computer vision applications such as image stitching, image retrieval and visual localization. However, it is often affected by many practical factors which will degrade its performance. Among these factors, illumination variations are the most influential one, and especially no previous descriptor learning works focus on dealing with this problem. In this paper, we propose IF-Net, aimed to generate a robust and generic descriptor under crucial illumination changes conditions. We find out not only the kind of training data important but also the order it is presented. To this end, we investigate several dataset scheduling methods and propose a separation training scheme to improve the matching accuracy. Further, we propose a ROI loss and hard-positive mining strategy along with the training scheme, which can strengthen the ability of generated descriptor dealing with large illumination change conditions. We evaluate our approach on public patch matching benchmark and achieve the best results compared with several state-of-the-arts methods. To show the practicality, we further evaluate IF-Net on the task of visual localization under large illumination changes scenes, and achieves the best localization accuracy.

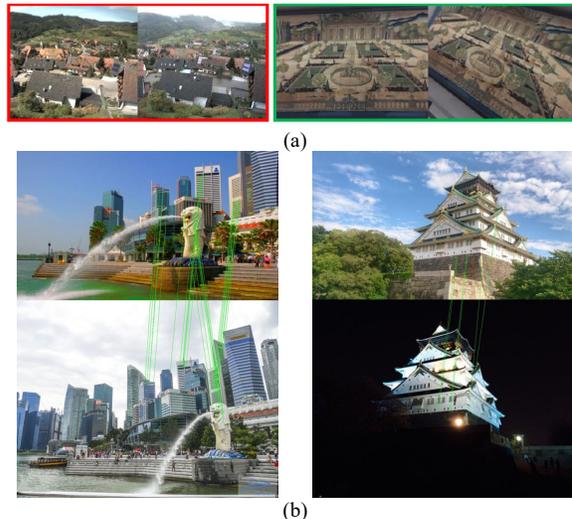

Figure 1: (a) Current benchmark focus on geometry-related scenes and with only small illumination variations. (b) Our work can tackle both geometry and illumination-related scenes by generating robust and generic descriptors.

## I. INTRODUCTION

Designing powerful local feature descriptor is a fundamental and critical problem in many computer vision applications such as image retrieval [25, 30], camera localization [15], wide baseline stereo [23], and structure-from-motion (SfM) [26, 27]. The representation of these local features must be capable to describe a variety of physical aspects, such as the variations in scale, viewpoint or illumination. These works can be divided into two categories: hand-crafted features and learned features using Convolutional Neural Network (CNN).

In hard-crafted features, the well-known SIFT [18] improves accuracy in feature matching, opening a new age in the field of feature description. [9] uses a binary method to generate a robust feature descriptor, while others [12, 13] explore to improve the performance by using an alternative distance metrics.

In learning-based features, it can be further divided to two different kinds based on whether or not the metric layer is used. With the metric layer, [5, 15] replace traditional hand-crafted methods to generate a more robust descriptor. However, the drawback of using a metric layer is it cannot perform the Nearest Neighbor Search (NNS). Since this kind of method usually treat the matching task only as a binary classification problem. On the other hand, the CNN architecture of [19, 20, 22, 31] do not contain any metric layer, and the results are not limited to binary classification which can make the matching accuracy better. To this end, CNN architecture without the use of metric layer has become the mainstream in learning-based feature descriptors.

Despite the recent notable achievements by using CNN, the performance of current state-of-the-art learning-based descriptors is found out to be somewhat limited on standard benchmarks [1]. As shown in Figure 1(a), current standard benchmark dataset is either focus on scenes with only viewpoint changes or only minor illumination changes. This phenomenon often limits the generalizable ability of current learned feature descriptors, since most of the descriptors are focused only on learning how to handle scenes with viewpoint, orientation and scale variance. However, when feature matching is used as the pre-step of many real-life applications such as image retrieval or camera localization, where it usually happens that two images are taken in different weathers or more extreme, one is taken during the day while the other is taken at night, matching image pairs with significant illumination diversity become an extremely important issue.

To make feature matching feasible under large illumination diversity, we propose IF-Net, an Illumination-invariant Feature Network. Since the quality of training data and training strategy are crucial for the success of descriptor learning, and in order to make IF-Net capable of overcoming the effect caused by illumination changes, we propose an AMOS-crop dataset, which is a patch-correspondence dataset with very large illumination variations. However, instead of directly merging AMOS-crop with commonly used datasets,

Po-Heng Chen is with the Department of Electronic Engineering, National Chiao Tung University, Hsinchu, 300, Taiwan.
The rest of the authors are with the Department of Computer Science, National Chiao Tung University, Hsinchu, 300, Taiwan (email of corresponding author*: kuanwen@cs.nctu.edu.tw)

we evaluate the influence of dataset schedules. Interestingly, we find out directly combining all data into the training process leads to an inferior result, and this observation is consistent with [10]. Due to this reason, a new dataset scheduling methods named separation training scheme is proposed. The proposed training scheme not only can make IF-Net handle viewpoint changes scenarios, but also can largely improve the matching accuracy under significant illumination diversity, as shown in Figure 1(b).

Along with the training scheme, a new training loss function named Reduce-Outlier-Inlier (ROI) loss is also designed to train IF-Net. The proposed ROI loss can enhance the ability of IF-Net to resist data noise, which can stably converge the network and generate robust descriptors. Furthermore, in order to make anchor patch not only have better robustness to negative patch, but also to positive patch, a hard-positive mining strategy is proposed. As mentioned before, current benchmark dataset [1] often focus on evaluating on either viewpoint or scale changes scenarios. To show the effect of IF-Net with the proposed training scheme and loss function under severe illumination changes, we further propose an Image Matching Dataset (IMD). IMD is an evaluation dataset contains 54 worldwide attraction scenarios, each of which is categorized as either easy or hard, based on the level of illumination and viewpoint changes. As the proposed IF-Net achieves the best results not only on public benchmark dataset [1], but also on the proposed IMD, we believe IF-Net can also perform well on the task of camera localization under severe illumination diversity. To verify the statement, we show IF-Net actually achieves state-of-the-art results on single-shot localization compared to previous work [22].

In summary, the contributions of this paper has four folds:

- High quality training data and strategy is crucial for the success of descriptor learning. We find out that not only the kind of data is important but also the order in which it is presented. To respond to this phenomenon, a new dataset scheduling method named separation training scheme is proposed, which can significantly improve the matching accuracy and descriptor generalizable ability.
- A new training loss function named ROI loss is designed along with the proposed separation training scheme. The ROI loss not only strengthen the matching accuracy on geometry-related scenes (e.g. Hpathces benchmark), but also make CNN more capable of handling scenes with large illumination diversity (e.g. proposed IMD).
- In order to make anchor patch not only have a better robustness to negative patch, but also to positive patch, a hard-positive mining strategy is proposed. This strategy can more effectively widen the descriptor distance between (anchor, positive) and (anchor, negative), improving the discriminative ability of the network.
- We present two datasets for both training and evaluation, named AMOS-crop dataset and Image Matching Dataset, respectively. AMOS-crop dataset is a large collection of corresponding feature patches under a variety of illumination conditions from fixed cameras. IMD contains 54 world attractions with large illumination diversity, which can be used as the baseline to evaluate the ability of illumination invariance. The proposed datasets are available at http://covis.cs.nctu.edu.tw/IF-Net/.

## II. RELATED WORKS

The works on designing feature descriptors has gradually moved from handcrafted to learning-based methods. For classical handcrafted feature descriptors, please refer to [6, 21] for a detail overview. Here, we foucs on the review of descriptor learning, ranging from traditional method to popular CNN-based methods.

**Traditional descriptor learning.** Early efforts to learn descriptors are not limited to any particular machine learning approach, thus many unique works are proposed at that time. In contrast to SIFT [18], PCA-SIFT [14] does not directly use smoothed weighted histograms to gradient patch, but replace it with Principal Components Analysis (PCA). To achieve better performance, Simonyan et al. [29] and Brown et al. [3] focus on the learning of pooling region and non-linear transforms with dimensionality reduction, respectively. Not only in the field of float descriptors, there are also some works focus on learned binary descriptors. To ensure each bit in the binary descriptor can achieve low variance for intra-class and high variance for inter-class, BOLD [2] is proposed for adaptive online selection of binary intensity tests. Another work of binary descriptors is RMGD [7]. They combine an extended Adaboost bit selection with the proposed spatial ring-region based pooling method for intensity tests. All the mentioned works have one thing in common, which is the descriptor starts learning from low level features such as patch gradient or binary intensity test. Learning from low level features will make these works suffer from information loss.

**CNN based descriptor learning.** At the beginning, Siamese network is the main stream and the most commonly used architecture in CNN based descriptor, and a fully connected layer is usually used as the metric network to improved performance. MatchNet [8] is a typical Siamese network with metric layer. They show a great potential of CNN based descriptor learning by significantly improve previous results. Based on MatchNet, Zagoruyko et al. [33] attempt different kinds of network structure and propose a central-surround method to boost matching accuracy. By proposing a global loss function, and combine with metric layer, Kumar et al. [17] achieve the best performance on Brown dataset [3]. Despite metric network improves the matching accuracy, it also limits the versatility of the deep network (i.e. cannot perform nearest neighbor search).

To tackle this problem, later proposed CNN-based descriptors do not adopt metric layer into the model architecture. Simo-Serra et al. propose DeepDesc [28], a CNN without metric layer with aggressive mining strategy to select hard patches. L2-Net [31] yields significant improvements by modifying CNN architecture with proposed novel loss functions and progressive sampling strategy. Built on the basis of L2-Net, Mishchuk et al. [22] propose HardNet, a CNN network trained with a distance matrix to find hard-negative pairs and achieve state-of-the-art performance on public benchmark dataset [1]. Recently, Luo et al. additionally consider image geometry information and cross-modality contextual property into training process, and propose GeoDesc [20] and ContextDesc [19], respectively. However, both of the work not only consider patch as the network input, but also take the keypoint position information into account to train the descriptors, which is different from our approach and [22, 31]. Though mentioned works lead CNN-based descriptor to achieve significant improvements, they mainly focus on generating viewpoint-invariance and scale-invariance descriptors. However, illumination-invariance is also a non-negligible effect in the field of feature matching, but none of them take this effect into account. So our work aim to generate a descriptor meets both illumination-invariance and viewpoint-invariance, which seeks to achieve improvements not only in feature matching, but also in real-life applications such as camera localization.

Table 1. Results of difference dataset scheduling method evaluated on HPatches benchmark [1]. With the proposed separation training scheme, IF-Net achieves the best mean average precision (mAP) on the matching task.

| Dataset | $S_{basic}$ | $S_{fine}$ | $S_{separation}$ |
|---|---|---|---|
| Mixed | 0.31 | - | - |
| PhotoSynth | 0.45 | - | - |
| AMOS-crop | 0.21 | - | - |
| PS→AMOS-crop | - | 0.29 | - |
| AMOS-crop→PS | - | 0.34 | - |
| Separation | - | - | **0.48** |

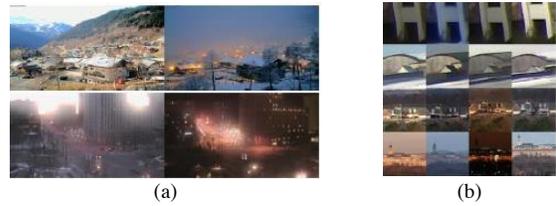

Figure 2. (a) Original AMOS dataset. Each row represents the same scene but during day and night time respectively. (b) Proposed AMOS-crop dataset. Each row corresponds to the same 3D point but with different illumination conditions.

### III. DATASET SCHEDULING

The quality of training data is crucial for the success of supervised learning. In this section, we first briefly introduce the proposed AMOS-crop dataset, and then investigate the matching performance depending on the presented training data. Interestingly, we find out the outcome is consistence with Ilg et al. [10]: *During the training process, not only the kind of training data will influence the results, but also the order in which it is presented*.

#### A. The AMOS-crop Dataset

The PhotoSynth (PS) dataset [24] is currently the largest training dataset for feature matching. It includes more than six million patch correspondences, and is collected automatically based on SIFT and structure-from-motion (SfM) technique. Limited to their process for dataset construction, PS dataset can only collect patches with little illumination diversity. However, the quality of training data will directly influence the performance of network in supervised learning. In order to collect patch correspondences with large illumination diversity, we take advantage of AMOS dataset [11] (Figure 2(a)). AMOS dataset is a collection of time-lapse photographs, containing images taken by fixed cameras at day and night for several months. Due to the fixed camera property, we construct AMOS-crop dataset, consisting by large amounts of feature correspondences under different weather and illumination conditions. To ensure the diversity of AMOS-crop dataset, we select 200 frames in each scene. To detect feature points in every frame, we apply a commonly used and stable detectors such as SIFT, SURF and ORB. By leveraging the property of fixed camera position, we can easily verify whether the detected features are the same or not by their pixel coordinates. Finally, as shown in Figure 2(b), we generate 1,380,724 patch correspondences with size 64x64 in 35 different scenes.

#### B. Dataset Scheduling Method

We adopt the network architecture introduced by Tian et al. [31] as our backbone model for testing different dataset schedules. The model is trained on both PS dataset and proposed AMOS-crop dataset. However, instead of directly merge both datasets for training, we investigate various dataset scheduling methods by considering different dataset training order and training mechanisms.

To train IF-Net on different datasets, we investigate 3 different training schedules. $S_{basic}$ is the basic schedule to train IF-Net for 50 epochs. Apart from $S_{basic}$, a schedule for fine-tuning $S_{fine}$ is also investigate. The last one $S_{saperation}$ is carefully designed relative to the first two schedules. In $S_{saperation}$, it first input PS and AMOS-crop datasets into a shared-weight IF-Net (Figure 3), and then update the parameters of the model by considering the gradients caused by both datasets. Different schedules and their results on HPatches benchmark are listed in Table 1. The results lead to the following observations:

**The order of training data with different properties matters.** Though AMOS-crop dataset provides scenarios with illumination diversity, however, merging with PS dataset to train IF-Net lead to worse results. Even for the schedules with fine-tuning, the performance is still not as expected. The results indicate the importance of training data schedules when learning a generic concept with deep networks.

**Separation training scheme outperforms mixed all data together.** By just using the proposed separation training scheme, we can improve IF-Net by almost 10% on HPatches benchmark. We conjecture that in order to let both datasets contribute equally influence to IF-Net, it's critical to first separate their descriptor generation process, then update the network by considering both gradients until back-propagation. Besides, we did not use specialized training sets for special scenarios. The above observation is in consistent with [10], where the dataset schedules are important for training a deep model. In the rest of the paper, we adopt IF-Net trained under $S_{saperate}$ as our default structure since it achieves the best results on public benchmark.

### IV. ILLUMINATION-INVARIANT FEATURE NETWORK

In this section, we will introduce the proposed methods for training IF-Net. Figure 3 shows the model architecture of deep network.

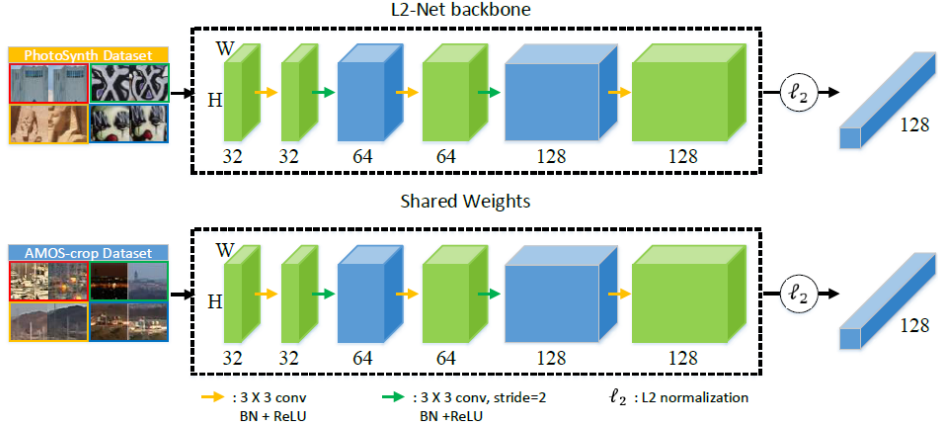

Figure 3. Proposed IF-Net architecture, L2-Net [31] is adopted as the backbone. By applying the proposed separation training scheme, two different kinds of training dataset (e.g. PS-dataset [24] and AMOS-crop dataset) will serve as the inputs to the shared-weight IF-Net and generate a 128-dims descriptor. Both kinds of the descriptors are normalized in order to reduce the impact within different sampled batch in every iteration.

## A. Hard-Positive Mining

The conventional triplet margin loss is used to increase the distance between matching and non-matching pairs, as shown in Eq. (1):

$$L = max(m + d(a,p) - d(a,n), 0), \quad (1)$$

where $d$ is the Euclidean distance, $a$, $p$ and $n$ represent anchor, positive and negative patch, respectively. Here positive means correct matching pair (i.e. correspondence) with anchor, while negative means the opposite. $m$ is a self-defined threshold to control the distance between positive and negative patches.

As most of the existing datasets [1, 24] belong to geometry-related samples, current work [22] only need to apply hard-negative mining to search hard-negative patches for training. However, for large illumination variation, considering only hard-negative patches is not enough. As shown in Figure 2, some matching pairs may appear quite different. How to sample a robust matching pair to improve network's discriminative ability becomes a critical and inevitable issue.

To solve this issue, we propose hard-positive mining strategy to strengthen network's ability. At First, a batch of matching pairs $X = \{(A_i, P_i^j) | i = 1..N, j = 1..M\}$ are generated, where $A$, $P$ represent anchor and positive patch respectively. $A_i$ and $P_i^j$ are the 2D points from different images $j$ but represents the same 3D point $i$ in the world (i.e., a matching pair). $M$ is the number of randomly sampled positive patch corresponding to the anchor. Then, $N \times (M+1)$ patches are forwarded to IF-Net to generate corresponding descriptors. In order to perform hard-positive mining strategy, an L2 pairwise distance matrix $D$ of size $N \times M$ is calculated. Each row in $D$ represents the L2 distance between one anchor patch and all of its corresponding matching pairs. In the process of hard-positive mining, we eventually select a positive patch with the farthest distance to the anchor, and serves as the correspondence matching pairs. As for the negative patch (i.e. non-matching pairs to the anchor), we follow [22] for the same sampling strategy.

By adopting the proposed hard-positive mining, the triplet margin loss from Eq. (1) are modified as Eq. (2):

$$L = \frac{1}{n}\sum_{i=1}^{n} max\,(m + max\,(d(a_i, p_i^j)) \\ - min\,(d(a_i, n_k), d(n_k, i_i^j)), 0), \quad (2)$$

where $a$, $p$ and $n$ share the same definition as Eq. (1), however, with additional subscription for hard-positive mining. In the subscript, $i$ represents $i$'th training data in a batch $X$ and $k \in [1, n], k \neq i$ for both $d(a_i, n_k)$ and $d(n_k, p_i^j)$. After the mining strategy, $n$ triplet training data $(a_i, p_i^j, n_k)$ is formed to train the network. With the proposed hard-positive mining, the network not only can make negative patch further away from anchor, but also can make hard-positive patch be closer to anchor.

## B. Reduce-Outlier-Inlier (ROI) Loss

Adopting the proposed mining strategy can strengthen the discriminative ability of the network. However, there still exist discrepancy between the training data in the distribution of descriptor space. This phenomenon will make the training process difficult to converge to a local minima value. To address this issue, we further propose a Reduce-Outlier-Inlier (ROI) loss, which can make the training process more stable with improved matching results. In the following, we assume all data have already been applied hard-positive mining and the strategy in [22]. Besides, $d_M(a_i, p_i)$ and $d_m(a_i, n_i)$ are denoted as the distance of the farthest matching pairs and the closest non-matching pairs, where $(a_i, p_i, n_i)$ is a batch of training data. As the above notations, the proposed ROI loss is formulated as Eq. (3):

$$L = \frac{1}{n}\sum_{i=1}^{n} max\,(m + \sigma\left(\frac{1}{n}\sum_{i=1}^{n} d_M(a_i, p_i)\right) d_M(a_i, p_i) \\ - \sigma\left(\frac{1}{n}\sum_{i=1}^{n} d_m(a_i, n_i)\right) d_m(a_i, n_i), 0) \quad (3)$$

Intuitively, the ROI loss seeks to increase the distinctiveness of descriptors by penalizing the effect caused by any outlier patches. Since in a forward batch of training data, there are usually some patches whose distance (in descriptor space) are far from the average. It is a critical and inevitable issue especially in illumination-related data (Amos-crop). Thus, the proposed ROI loss takes into account the data distribution as the weighting score of the distance value in each of the forward batch. In order for the loss to be minimized, the outlier matching and non-matching pairs will get higher scores, encourage the loss to be minimized and vice-versa – inliers will lead to similar scores, so there is no need to additionally consider the weight within a batch of data. The proposed ROI loss can be served as a plug-in of triplet margin loss by re-formulating as the weighted-average version:

$$L = \frac{1}{n}\sum_{i=1}^{n} \max(m + w_p d_M(a_i, p_i) - w_n d_m(a_i, n_i), 0). \quad (4)$$

Table 2. Mean average precision on HPatches benchmark [1]. IF-Net (Ours) performs the best on both matching and retrieval tasks. As for verification task, IF-Net only behind the best results in a very small and acceptable margins (0.02 for *Inter* and 0.04 for *Intra*).

| Training Dataset | | None | PhotoSynth (PS) Dataset | | | PS Dataset + AMOS-crop Dataset | |
|---|---|---|---|---|---|---|---|
| | | SIFT | DeepDesc [28] | DC-2-Str [33] | Hard-Net | Hard-Net [22] | Ours |
| Verification | *Inter* | 82.83 | 92.12 | 91.49 | 93.33 | **93.85** | 93.83 |
| | *Intra* | 80.35 | 89.12 | 89.07 | **93.61** | 93.53 | 93.57 |
| Matching | | 24.44 | 25.60 | 25.50 | 44.76 | 41.64 | **48.65** |
| Retrieval | | 41.71 | 52.71 | 45.77 | 63.52 | 60.85 | **67.31** |

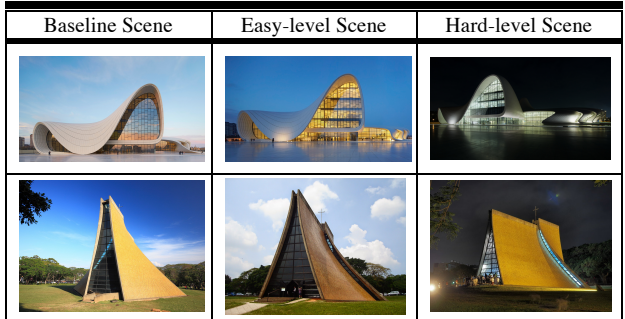

Figure 4. Two examples from the proposed Illumination Matching Dataset (IMD). Easy-level scene contains only minor illumination changes while hard-level scene includes severe diversities on both illumination and viewpoint aspects.

## V. EXPERIMENTS

### A. Implementation

**Training details.** The proposed IF-Net is trained for 50 epochs using Adam [16] with an initial learning rate of 0.1, which was further divided by 10 every 10 epochs after 30th epoch. The batch size is set to 512, and to extract patches from raw images, TILDE [32] and SIFT are used as the feature detector. And the generated descriptors are normalized by applying L2 normalization.

**Training dataset.** PhotoSynth (PS) dataset [24] is widely used for learning local image descriptors, it consists of diversity scenes with viewpoint, scales in comparison to the MVS dataset [3]. Based on this reason, PS-dataset is chosen to serve as the geometry-related dataset to train IF-Net. As for illumination-related dataset, the proposed AMOS-crop dataset is used for training.

### B. Evaluation Datasets

**Homography dataset.** HPatches [1] is a large-scale patch dataset for evaluating local features. It consists of large viewpoint and small illumination changes. As HPatches provides groundtruth homogrphies and raw images, it can also be used to evaluate image matching performance. For evaluation, we refer to [1], consisting of 116 sequences and 580 image pairs.

**Illumination Matching Dataset.** Though HPatches dataset consists data with illumination change, the coverage of the diversity does not enough to represent real-life scenarios. To this end, we propose an Illumination Matching Dataset (IMD), consisted of 54 world attraction scenes selected from Google Image search. Each attraction scene contains an easy-level and hard-level pair, and the level is defined by its illumination and viewpoint diversity. Some examples are shown in Figure 4. It can be clearly seen that IMD contains many significant illumination and viewpoint changes scenarios, which are quite difficult for feature matching.

**NTU Localization Dataset.** To demonstrate the generalization of the proposed IF-Net, we further evaluate the localization ability of our work under severe illumination diversity. We use NTU long-term positioning dataset [4] for evaluation.

### C. Evaluation Protocols

**Patch level.** We follow HPatches [1] evaluation protocols and use mean average precision (mAP) for three sub-tasks, including patch matching, retrieval and verification.

**Image level.** To extract keypoints from IMD raw images, we apply TILDE [32] as the feature detector on HardNet and IF-Net. Since the proposed IMD contains significant illumination and viewpoint changes, it's hard to obtain a unified and robust ground truth correspondences. So in the following matching accuracy calculation, we manually verify the results on both IF-Net and HardNet.

**Localization level.** The localization is evaluated under different sessions in different lighting conditions (by different weather). 288 images from one session (M1/20 15:00) are used for model reconstruction, which follows the evaluation protocol in [4], and each test session includes 108 images with ground truth locations. Since localization accuracy and image registration rate are trade-off relations, thus we list both of them for fair comparison.

### D. Evaluation Results

**HPatches Evaluation.** To demonstrate the efficacy of proposed training schemes of IF-Net, we evaluate several works on HPatches benchmark, including verification, matching and retrieval. While HardNet is the most relevant w ork to ours, we also train HardNet on both PS and AMOS-crop datasets for fair comparison. As shown in Table 2, the proposed training schemes of IF-Net improves the overall performance over the previous best-performing under similar training settings. Furthermore, in both image retrieval and matching tasks, our approach leads significant accuracy improvement against Hard-Net, which demonstrate the robustness of descriptor generated from IF-Net.

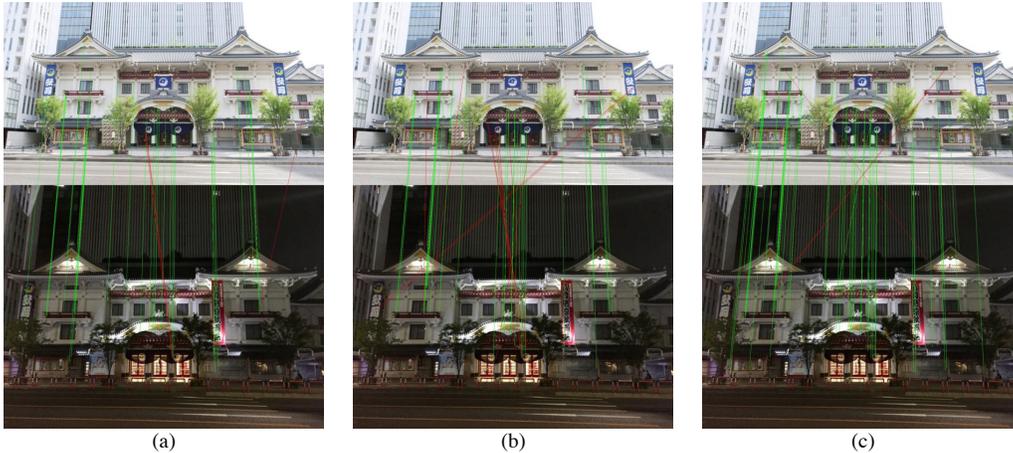

(a)                                   (b)                                   (c)

Figure 5. Comparison between proposed IF-Net and HardNet [22]. (a) IF-Net. (b) HardNet (PS + AMOS dataset). (c) HardNet (PS dataset). As can be seen clearly, our method outperforms state-of-the-art under the hard-level scene by matching more correct pairs and less incorrect pairs.

Table 3. Precision rate comparison on proposed IMD.

|  | Scene level | HardNet (Trained on PS+AMOS) | HardNet (Trained on PS) | IF-Net |
|---|---|---|---|---|
| Total | Easy | 70.9% | 76.9% | **79.7%** |
|  | Hard | 55.3% | 61.1% | **62.4%** |
| Top-40 | Easy | 86.6% | 92.5% | **94.2%** |
|  | Hard | 72.7% | 80.7% | **83.2%** |

**IMD Evaluation.** To show how the proposed IF-Net can handle practical scenarios of image matching, we evaluate IF-Net and HardNet on the proposed IMD. To give a fair comparison, HardNet has two different versions: one is trained on only PS-dataset and the other is trained on both PS and AMOS-crop dataset. The qualitative and quantitative results are shown in Table 3 and Figure 5, respectively. We report two different kinds matching accuracy: *Top-40* and *Total*. The reason to do report Top-40 is due to some scenarios in IMD contains too much illumination and viewpoint changes, and all methods perform poorly on such cases, which will lead to a non-representative evaluation. As can be seen from Table 3, IF-Net achieves the best performance regardless of which accuracy scheme is considered. This demonstrate our approach can still generate robust and generic feature descriptors under scenes with large illumination changes.

**Localization.** To justify our approach can tackle a more challenging condition, we evaluate IF-Net on a complex computer vision task: visual localization. The evaluated results are in Table 4, except for the registration rate, the lower value implies better performance. Since both SIFT and HardNet trained on PS and AMOS have significant lower registration rate than ours (i.e. shows inferior performance on visual localization), we will not consider to compare the localization error between these methods. Instead, as can be seen from Table 4, our approach achieves the best results on both registration rate and localization accuracy under almost all lighting conditions compared to HardNet trained on PS dataset. This shows IF-Net can truly generate illumination-invariance descriptors.

Table 4. Localization error (cm) and registration rate (R-rate) on NTU-dataset [4].

| Test date & time |  | M2/3 10:30 | M2/5 15:00 | M2/6 12:00 | M2/7 14:00 | M2/10 10:00 |
|---|---|---|---|---|---|---|
| Weather |  | Cloudy | Sunny | Sunny | Cloudy | Cloudy |
| SIFT [18] | Mean | 0.853 | 0.760 | 0.704 | 0.591 | 0.776 |
|  | Stdev. | 1.090 | 1.265 | 0.622 | 0.410 | 0.881 |
|  | R-rate | 72.2% | 74.0% | 69.4% | 77.7% | 66.6% |
| HardNet [22] (Trained on PS+AMOS) | Mean | 0.217 | 0.187 | 0.243 | 0.208 | 0.229 |
|  | Stdev. | 0.191 | 0.081 | 0.233 | 0.116 | 0.165 |
|  | R-rate | 87.0% | 91.6% | 66.6% | 75.0% | 78.7% |
| HardNet (Trained on PS) | Mean | 0.326 | 0.325 | 0.381 | 0.380 | 0.324 |
|  | Stdev. | 0.130 | 0.135 | 0.319 | 0.337 | 0.126 |
|  | R-rate | 100% | 100% | 99.0% | 96.2% | 100% |
| IF-Net | Mean | 0.310 | 0.309 | 0.337 | 0.327 | 0.314 |
|  | Stdev. | 0.129 | 0.132 | 0.156 | 0.140 | 0.150 |
|  | R-rate | **100%** | **100%** | 98.1% | 98.1% | **100%** |

## VI. CONCLUSIONS

In this paper, we propose an Illumination-invariant Feature Network (IF-Net) to tackle image matching problems under large illumination diversity. IF-Net is trained with the proposed ROI loss function and separation training scheme, the former can make deep model be less affected by the noise of training data while the later can balance the distribution between different training data. Furthermore, we construct an AMOS-crop dataset for training and an IMD dataset for matching evaluation. We evaluate IF-Net on both HPatches matching benchmark and NTU visual localization task. The results show that IF-Net achieves state-of-the-art performance, which proves the efficiency and practicality of the proposed method.

## VII. ACKNOWLEDGEMENT

This research was supported in part by the Ministry of Science and Technology of Taiwan (MOST 108-2633-E-002-001, MOST 107-2221-E-009-148-MY2, and MOST 109-2634-F-009-015-), National Taiwan University (NTU-108L104039), and Delta Electronics.